# Détection automatique des infections du site opératoire à partir d'un entrepôt de données

Marine Quéroué[a], Agnès Lashéras-Bauduin[b], Vianney Jouhet[ac], Frantz Thiessard[ac],
Jean-Marc Vital[d], Anne-Marie Rogues[be], Sébastien Cossin[ac]

[a] *CHU de Bordeaux, Pôle de Santé Publique, Service d'Information Médicale, Informatique et Archivistique Médicales (IAM), Bordeaux, F-33000, France*
[b] *CHU de Bordeaux, Pôle de Santé Publique, Service d'Hygiène Hospitalière, Bordeaux, F-33000, France*
[c] *Université de Bordeaux, Bordeaux Population Health Research Center, équipe ERIAS, UMR 1219, F-33000 Bordeaux, France*
[d] *CHU de Bordeaux, Pôle de chirurgie, Service de chirurgie orthopédique et traumatologique, unité de chirurgie du rachis, Bordeaux, F-33000, France*
[e] *Université de Bordeaux, Bordeaux Population Health Research Center, ISPED, UMR 1219, F-33000 Bordeaux, France*

**Résumé**

*La réduction de l'incidence des infections du site opératoire (infection associée aux soins survenant à la suite d'un acte de chirurgie) fait partie des objectifs du programme national de lutte contre les infections nosocomiales. Pour ce faire une surveillance manuelle est réalisée chaque année par l'équipe d'hygiène hospitalière et les chirurgiens du CHU de Bordeaux. Notre objectif était de développer un algorithme de détection automatique des ISO à partir des données du système d'information hospitalier. Les fiches de surveillance de la chirurgie du rachis des années 2015, 2016 et 2017 ont servi de gold standard pour classifier les actes de chirurgie et entraîner des algorithmes d'apprentissage automatique. Notre jeu d'apprentissage comprenait 22 ISO parmi 2133 actes de chirurgie du rachis. Des variables prédictives d'ISO ont été extraites à partir de l'entrepôt de données i2b2. Nous avons comparé deux approches différentes. La première utilise plusieurs sources de données et offre les meilleures performances mais est difficilement généralisable à d'autres établissements. La seconde est basée uniquement sur le texte libre avec extraction semi-automatique des termes discriminants puis apprentissage automatique. Les algorithmes réussissent à identifier l'ensemble des ISO avec 20 et 26 faux positifs respectivement sur le jeu d'apprentissage. Une évaluation sur des nouvelles données est en cours. Ces résultats sont encourageants pour le développement de méthodes de surveillance semi-automatisée.*

*Mots-clés:*
Surgical Wound Infection
Datawarehouse
Natural Language Processing

## Introduction

Entre 2012 et 2017, la proportion d'infections du site opératoire (ISO) est passée de 13,5 % à 15,92 % des infections associées aux soins (IAS) les classant en deuxième position après les infections urinaires d'après l'enquête nationale de prévalence des infections nosocomiales [1]. Durant cette période, les proportions d'ISO profondes et au niveau de l'organe ont augmenté (respectivement 4,8 % vs 5,77 % et 5,5 % vs 7,74 %). Seules les ISO superficielles ont vu leur proportion baisser (3,2 % vs 2,41 %). Dans le rapport 2017 de surveillance des ISO dans les établissements de santé, les taux d'incidence étaient en 2017 de 1,37 % pour la chirurgie orthopédique, 1,97 % pour la chirurgie digestive, 1,88 % pour la gynécologie-obstétrique, 1,10 % pour la traumatologie, 2,60 % pour l'urologie, 0,79 % pour la neurochirurgie, 1,72 % pour la chirurgie bariatrique, 3,44 % pour la chirurgie coronaire, 3,99 % pour la chirurgie réparatrice et reconstructive, 1,32 % pour la chirurgie thoracique et 2,32 % pour la chirurgie vasculaire [2].

Améliorer la surveillance et la prévention des ISO fait partie de l'axe 3 du programme national d'actions de prévention des IAS (Propias) [3]. Il a pour objectif de « disposer d'outils de surveillance des ISO graves (profondes ou nécessitant une reprise chirurgicale), d'évaluation de leur prévention et de gestion adaptés dans les 3 secteurs de l'offre de soins ».

Jusqu'en 2018, la surveillance des ISO en France était proposée aux établissements de santé volontaires par le réseau ISO-Raisin selon 2 modalités. La surveillance dite prioritaire concerne une liste d'interventions sentinelles. Il s'agit d'un recueil d'informations avec création de fiches concernant au moins 100 interventions consécutives de la même spécialité pendant les 6 premiers mois de l'année. La surveillance dite globale ou agrégée se fait quant à elle sur une période d'au moins deux mois au cours du premier semestre, concerne des interventions incluses ou non dans la liste prioritaire et nécessite la création de fiches seulement pour les ISO déclarées [2]. Cette surveillance s'avère être extrêmement chronophage du fait du recueil de données qu'elle impose et de la validation des ISO par les chirurgiens.

En novembre 2018, le CPias Ile-de-France a été nommé par Santé Publique France pour le pilotage de la mission nationale « Surveillance et prévention du risque infectieux liés aux actes de chirurgie et de médecine interventionnelle » (Spicmi). Cette mission a pour vocation le remplacement du réseau actuel ISO-Raisin [4]. Elle a notamment pour objectif de passer à un nouveau système de surveillance reposant principalement sur l'utilisation des données des systèmes d'information hospitalier (SIH) des établissements de santé. Cette surveillance « semi-automatisée » représenterait un gain en termes de ressources nécessaires pour la collecte des données dans chaque établissement.

Plusieurs structures ont déjà étudié la possibilité d'automatiser la surveillance des IAS et notamment des ISO. C'est le cas du CHU de Nantes qui a évalué l'utilisation des données du programme de médicalisation des systèmes d'information (PMSI) et de la bactériologie pour détecter les ISO [5]. Selon eux, une surveillance assistée par ordinateur peut être mise en œuvre dans les hôpitaux français en utilisant des sources de données disponibles. Le gain de temps d'une détection semi-automatisée permettra aux professionnels de la prévention des infections associées aux soins de consacrer plus de temps aux tâches de prévention et d'éducation. Ils soulignaient cependant la nécessité d'une étude multicentrique pour évaluer la transposabilité de cette méthode.

En dehors de la France des recherches sont également menées dans ce sens. L'Université de Stockholm a étudié l'utilisation de techniques d'apprentissage automatique pour la détection des infections associées aux soins. Les chercheurs ont utilisé des données hospitalières recueillies lors d'une enquête de prévalence ponctuelle. Elles comprenaient des données textuelles, des codes CIM10 (classification internationale des maladies, $10^{ème}$ révision), des administrations médicamenteuses, des résultats microbiologiques et la température corporelle. Les algorithmes d'apprentissage séparateurs à vaste marge (SVM) et gradient tree boosting (GTB) se sont révélés performants pour cette tâche avec un excellent rappel [6].

Dans le cadre du projet national d'amélioration de la qualité chirurgicale (NSQIP) aux États-Unis, l'Université du Minnesota a également mis en place un outil automatisé de détection des événements indésirables liés à une intervention chirurgicale. Ces derniers ont utilisé les données cliniques de leur centre médical ainsi que celles du registre NSQIP. Parmi celles-ci, ils ont sélectionné des données démographiques (sexe, âge, race) et des données cliniques (diagnostics CIM-9, résultats biologiques, administrations médicamenteuses, demandes d'examens complémentaires et constantes). Les modèles utilisant la régression logistique ont montré les meilleures performances éliminant de manière fiable la grande majorité des patients sans ISO et réduisant ainsi de manière significative la charge des registres [7].

Enfin, l'Université de l'Utah a quant à elle développé un système de traitement automatique du langage naturel (TAL) pour identifier automatiquement les mentions d'ISO dans les comptes rendus de radiologie. Ils ont travaillé sur la chirurgie gastro-intestinale à partir de la base de données MIMICIII Critical Care dont ils ont extrait les codes diagnostiques, les codes d'acte et les comptes rendus de scanner dans les 30 jours suivant la procédure. Ces derniers ont été annotés par deux chirurgiens afin de créer un lexique de termes. Ils ont ensuite développé un système de TAL afin d'identifier et classer automatiquement les preuves d'ISO à partir de chaque compte rendu en s'appuyant sur une adaptation de l'algorithme ConText qui gère la négation et la temporalité. Leur système de TAL s'est montré plus performant que les deux autres approches testées utilisant des données administratives uniquement ou des techniques d'apprentissage automatique SVM avec une représentation du texte en n-grammes [8].

Le CHU de Bordeaux a mis en œuvre en novembre 2017 un entrepôt de données de santé basé sur la solution open source i2b2 [9] pour faciliter la réutilisation des données à des fins de recherche, de prévention et d'amélioration de la qualité des soins. L'un des cas d'usage de l'entrepôt est de développer des méthodes de détection automatique des ISO en collaboration avec le service d'hygiène hospitalière chargé de leur surveillance et de leur prévention. Cet article présente les premiers résultats de l'implémentation d'une méthode de détection automatique des ISO de la chirurgie du rachis au CHU de Bordeaux.

## Méthodologie

Dans le cadre de la surveillance nationale organisée par le réseau ISO-Raisin, et afin d'évaluer l'incidence des ISO au sein du CHU de Bordeaux, le service d'hygiène hospitalière réalise des enquêtes chaque année sur des chirurgies ciblées. C'est le cas de la chirurgie du rachis pour laquelle l'ensemble des actes réalisés sur une période de 3 mois consécutifs par an est étudié. Parmi une liste des actes extraits, il est demandé à chaque chirurgien de signaler lesquels ont conduit à une ISO. Pour chaque ISO déclarée, les membres du service d'hygiène hospitalière réalisent alors un retour au dossier patient informatisé et une analyse des causes pour dégager des mesures de prévention.

Les fiches de surveillance ainsi créées constituent un gold standard avec respectivement 13 ISO déclarées sur 662 actes, 7 sur 708 actes et 2 sur 763 actes pour les années 2015, 2016 et 2017. Au total, 2133 actes de chirurgie du rachis ont été classés par les chirurgiens ayant réalisé ces actes et 22 ISO ont été déclarées sur ces trois années.

L'entrepôt de données du CHU de Bordeaux contient des données structurées (diagnostics CIM10, actes de la Classification Commune des Actes Médicaux - CCAM, prescriptions et administrations médicamenteuses, biologie et bactériologie), semi-structurées (formulaires) et des données non structurées (documents en texte libre). Cet entrepôt est alimenté quotidiennement par les données du SIH. L'ensemble des données de notre cohorte est mobilisable pour développer des méthodes d'apprentissage automatique afin de prédire automatiquement des nouveaux cas d'ISO et ainsi mettre en place une surveillance semi-automatisée au CHU de Bordeaux.

Utiliser l'ensemble des données disponibles limite la transposabilité de l'algorithme à d'autres établissements. En effet, les établissements de santé ont chacun des applications différentes et certaines données du CHU de Bordeaux, par exemple les résultats de bactériologie, ne sont pas structurées de la même façon dans un autre établissement. Deux algorithmes ont été développés pour prendre en compte ce compromis entre transposabilité et quantité d'information, le premier utilise l'ensemble des données disponibles, il est donc spécifique au CHU de Bordeaux ; le second se limite aux données en texte libre et pourrait être transposé à d'autres établissements.

Les données PMSI pour la tarification à l'activité, les administrations médicamenteuses, les données de la bactériologie ainsi que l'ensemble du texte libre ont été utilisées par le premier algorithme. Les variables du modèle ont été sélectionnées manuellement à partir de connaissances expertes des ISO de la chirurgie du rachis. Concernant le PMSI, des codes CIM10 relatifs à une ISO et pertinents dans le cadre d'une chirurgie orthopédique du rachis ont été sélectionnés, ainsi qu'un acte CCAM de reprise post-ISO validé par les chirurgiens (tableau 1). Parmi les administrations médicamenteuses, seules les antibiothérapies ont été prises en compte (codes J01 et J04 de la classification ATC). Parmi les données de la bactériologie, des protocoles dont la demande était faite lors d'une suspicion d'ISO (plaie opératoire, pus profond, matériel orthopédique, biopsie ostéo-articulaire ou autre, liquide de lame ou redon) ont été sélectionnés.

*Tableau 1 – Codes CIM10 et codes CCAM en rapport avec une infection du site opératoire de la chirurgie orthopédique du rachis sélectionnés par l'expert*

| Code CIM10 | Libellé |
|---|---|
| T81.4 | Infection après un acte à visée diagnostique et thérapeutique, non classée ailleurs |
| T84.5 | Infection et réaction inflammatoire dues à une prothèse articulaire interne |
| T84.6 | Infection et réaction inflammatoire dues à un appareil de fixation interne [toute localisation] |
| T84.7 | Infection et réaction inflammatoire dues à d'autres prothèses, implants et greffes orthopédiques internes |

| Code CCAM | Libellé |
|---|---|
| AFPA001 | Mise à plat de lésion infectieuse péridurale rachidienne et/ou paravertébrale postopératoire [sepsis], par abord direct |

Enfin concernant le texte libre, 12 termes spécifiques à l'ISO ont été sélectionnés par un expert du domaine.

Ainsi pour chaque acte de notre cohorte, dans un délai de 90 jours après la date d'intervention (période de surveillance requise pour la chirurgie du rachis) nous avons relevé la présence ou non d'un diagnostic relatif à une ISO signalée par le chirurgien, la présence ou non d'un acte de reprise, la présence ou non d'une administration antibiotique, la présence ou non d'un protocole bactériologique en rapport avec une suspicion d'ISO et pour chaque terme d'intérêt sa fréquence dans le texte libre.

Le second algorithme était uniquement basé sur les données non structurées en texte libre de l'entrepôt de données, notamment les comptes rendus opératoires, de consultation et d'hospitalisation dans un délai de 90 jours après l'intervention. Contrairement au premier algorithme, la sélection des termes d'intérêt a commencé par une étape de sélection automatique de termes. Un étiquetage morpho-syntaxique était d'abord réalisé par le logiciel TreeTagger [10] puis les groupes nominaux étaient extraits à l'aide d'une expression régulière. Les formes lemmatisées de ces groupes nominaux correspondaient aux termes d'intérêt. Par exemple, dans la phrase « reprise chirurgicale pour infection du site opératoire » les termes « reprise, reprise chirurgicale, reprise chirurgicale pour infection, infection, infection du site, infection du site opératoire, site, site opératoire » étaient extraits automatiquement. Le code de cet outil est publié en open-source[1].

Un premier filtre a permis d'exclure les termes non fréquents chez les patients atteints d'ISO. Un seuil arbitraire de 20 % a été fixé, si un terme survenait chez moins de 5 patients atteints d'ISO parmi les 22, le terme était exclu. Un odds ratio (rapport de cotes), une mesure utilisée en classification automatique de textes pour quantifier le lien entre un terme et une catégorie [12], a été calculé pour chacun des termes. Un deuxième filtre excluait les termes qui pourraient être trop spécifiques au CHU de Bordeaux. Le chapitre d'un livre sur les infections postopératoires de la chirurgie du rachis [11] a été utilisé : les termes ont été extraits par la même méthode décrite plus haut et les termes des dossiers patients non présents dans cette référence ont été exclus. Finalement, les 20 termes les plus associés aux ISO ont été conservés après validation par un expert du domaine. Cette méthode de sélection semi-automatique des termes est résumée dans la figure 1. Dans la matrice fournie aux algorithmes d'apprentissage, chaque patient était une ligne et chaque terme une colonne. Si le terme était retrouvé dans les 90 jours après la date d'intervention, l'information était codée 1 et 0 sinon.

Deux algorithmes de classification ont été testés : la régression logistique et les forêts aléatoires. Pour la détection semi-automatique des ISO, il est souhaitable d'obtenir une parfaite sensibilité. Les algorithmes ont été évalués dans ce sens : ils ont été comparés en fonction de leur spécificité pour une sensibilité fixée à 100%. Pour obtenir cette sensibilité, le seuil de prédiction a été fixé à la plus faible probabilité prédite parmi nos cas d'ISO.

## Résultats

### Premier algorithme

La régression logistique offre les meilleurs résultats avec l'ensemble des 22 ISO détectées sur les 2133 actes pour 20 faux positifs (tableau 2).

*Tableau 2 – Prédictions de la régression logistique avec l'ensemble des données sur jeu d'apprentissage (années 2015, 2016 et 2017). M : ISO, T : prédiction*

| Algorithme 1 | M+ | M- | |
|---|---|---|---|
| **T+** | 22 | 20 | 42 |
| **T-** | 0 | 2091 | 2091 |
| | 22 | 2111 | 2133 |

La spécificité est de 99,05 %, la valeur prédictive positive de 52,38 % et l'exactitude de 99,06 %. Un retour au dossier a été effectué concernant les 20 faux positifs : 8 étaient des ISO non déclarées (dont 2 superficielles), 9 étaient des actes de reprise post-ISO et 3 n'étaient pas des ISO. Le modèle utilisant les forêts aléatoires présente de moins bonnes performances avec une précision de 20,18 % (87 faux positifs) pour une sensibilité à 100 %.

Enfin nous avons évalué les performances de la régression logistique en scindant notre échantillon de départ. L'apprentissage se faisant sur les années 2015 et 2016, nous avons testé le modèle sur l'année 2017 (Tableau 3). Les 2 ISO signalées cette année ont correctement été détectées ainsi que 4 faux positifs sur 763 actes étudiés (valeur prédictive positive : 33,3 %).

---

[1] https://github.com/scossin/CandidateTerm

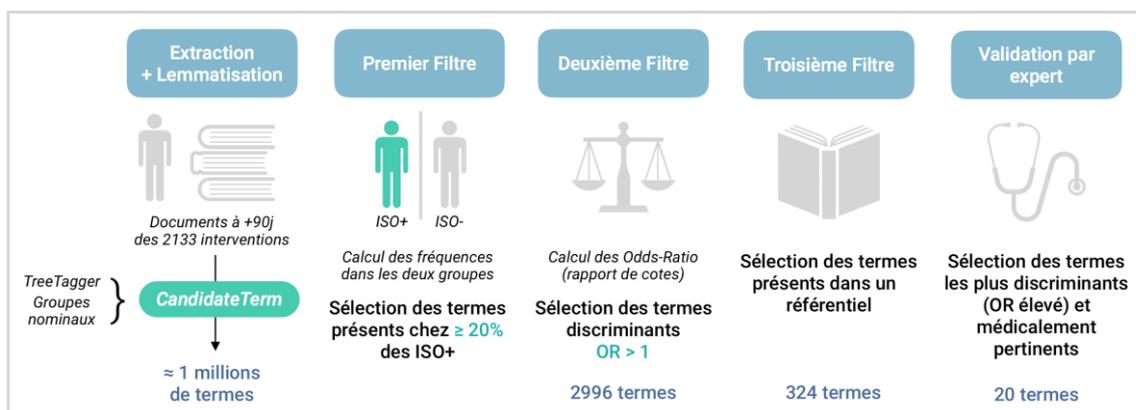

Figure 1 – Sélection semi-automatique des termes (algorithme 2). Un programme extrait automatiquement les termes (groupes nominaux) des documents. Les filtres conservent les termes fréquents chez les cas d'ISO, discriminants et spécifiques à la chirurgie du rachis. Enfin, un expert valide et conserve 20 termes cliniquement pertinents.

Tableau 3 – Prédictions de la régression logistique avec l'ensemble des données sur un jeu de test (année 2017) après apprentissage (année 2015 et 2016). M : ISO, T : prédiction

| Algorithme 1 | M+ | M- | |
|---|---|---|---|
| T+ | 2 | 4 | 6 |
| T- | 0 | 757 | 757 |
| | 2 | 761 | 763 |

Tableau 4 – Prédictions de la régression logistique avec données textuelles uniquement, sur jeu d'apprentissage (années 2015, 2016 et 2017). M : ISO, T : prédiction

| Algorithme 2 | M+ | M- | |
|---|---|---|---|
| T+ | 22 | 26 | 48 |
| T- | 0 | 2085 | 2085 |
| | 22 | 2111 | 2133 |

**Deuxième algorithme**

Environ un million de termes ont été extraits du texte libre et 2 996 termes conservés après le premier filtre qui excluait les termes non fréquents chez les patients atteints. Le terme « site opératoire » est survenu chez la totalité des patients atteints d'ISO et seulement chez 5 % des patients non atteints. Ce terme avait l'odds ratio (OR) le plus élevé.

Parmi ces 2 996 termes, seulement 324 termes ont été retrouvés dans l'ouvrage de référence sur les ISO de la chirurgie du rachis et ont donc été conservés par le deuxième filtre. Parmi les termes exclus, le terme « code AFPA001 » avait un OR très élevé (1302). Ce code d'acte CCAM a pour libellé « mise à plat de lésion infectieuses péridurale rachidienne et/ou paravertébrale postopératoire ». La présence de ce code, notamment dans les comptes rendus opératoires, est probablement trop spécifique au CHU de Bordeaux car les codes CCAM sont systématiquement présents.

Les 20 termes les plus associés aux ISO en termes d'odds ratio ont été conservés après validation par l'expert, dans l'ordre : site opératoire, antibiothérapie, sepsis, écoulement, parage, désunion de la cicatrice, lésion, plaie, infectiologues, lavage, écoulement purulent, rifadine, cicatrice, fils, staphylocoque, CRP, niveau de la cicatrice, rifampicine, point et antibiotique.

La régression logistique offre les meilleurs résultats et détectait l'ensemble des cas d'ISO au prix de 26 faux positifs (tableau 4).

## Discussion

**Résultats**

Ces premiers résultats, permettant de repérer l'ensemble des ISO pour très peu de faux positifs, sont très encourageants pour le développement de méthodes semi-automatiques.

Le premier algorithme utilise l'ensemble des données disponibles et est très spécifique aux données du CHU de Bordeaux. Sa meilleure capacité à séparer les ISO et les non ISO par rapport au deuxième algorithme utilisant uniquement le texte libre peut s'expliquer par le nombre élevé de sources de données utilisées : codes diagnostics, code d'acte, prélèvements bactériologiques, antibiothérapie et mention de l'ISO dans le texte libre. Utiliser plusieurs sources de données augmente la probabilité de trouver de l'information concernant une ISO chez un patient.

Bien que les résultats de cet article ne soient pas comparables avec d'autres travaux, les performances ont un ordre de grandeur similaire. L'algorithme développé il y a quelques années à Nantes reposant uniquement sur le PMSI et la bactériologie [5] a obtenu un rappel de 90 % avec une précision de 20 %. L'algorithme utilisant le TAL développé par l'Université de l'Utah présente quant à lui de très bonnes performances avec un rappel de 93 % pour une précision de 82 %.

Notre deuxième algorithme présente aussi des résultats intéressants. Il démontre que l'information est présente dans le texte libre pour la détection des ISO. Les termes détectés automatiquement dans le texte sont pertinents pour la détection des ISO et offrent des résultats similaires à une sélection manuelle des variables par un expert du domaine. Cette approche de sélection de variables pourrait être réutilisée pour la détection des ISO des autres chirurgies. La méthode proposée utilise une approche linguistique pour l'extraction de groupes nominaux. Une mesure statistique et des ressources externes sont utilisées pour sélectionner et filtrer les termes associés aux ISO. Dans un souci de transposabilité, filtrer les termes à partir d'une ressource externe est une étape importante pour retirer des termes trop spécifiques à notre établissement bien que ceux-ci pourraient améliorer la prédiction. Cet algorithme pré-entraîné au CHU de Bordeaux présente l'avantage d'être plus facilement transposable à d'autres établissements.

### Limites

Les performances des algorithmes sont surévaluées en l'absence d'un échantillon d'évaluation. Bien que fixée à 100 %, la sensibilité nécessite d'être mesurée sur un nouveau jeu de données.

La principale limite de cette étude est le petit nombre de cas (22) d'ISO dans le gold-standard et la qualité de ce dernier. Une enquête des faux positifs a permis de montrer des erreurs dans le gold-standard qui étaient soupçonnées. Ce premier modèle avec une parfaite sensibilité a été privilégiée pour détecter l'ensemble des ISO et explorer les faux-positifs. Certains cas d'absence d'ISO étaient en réalité des ISO non déclarées par les chirurgiens probablement à cause d'un biais de mémorisation (oubli du clinicien).

Une limite du premier algorithme est la sélection manuelle des variables d'intérêt par un expert du domaine, celle-ci est relativement chronophage et reste très spécifique à la chirurgie du rachis. Aussi un seul expert a procédé à la sélection. L'algorithme 2 est plus facilement généralisable car les étapes manuelles sont moins chronophages. Il serait intéressant d'automatiser la tâche de sélection des variables cependant valider manuellement la pertinence clinique des variables sélectionnées favorise l'acceptabilité de l'algorithme par les chirurgiens.

Une autre limite de ce travail est l'absence d'évaluation formelle de l'outil d'extraction automatique des termes des documents. Celui-ci est spécifique au français et pourrait être comparé aux outils similaires comme BioTex [12] et TermSuite [13].

Aucune détection de la négation ou de la temporalité n'est pour l'instant réalisée bien que des algorithmes en français existent [14,15].

### Perspectives

Ce premier algorithme va permettre de détecter automatiquement des nouveaux cas d'ISO au-delà de la période d'étude (3 mois pendant 3 années). Ces nouveaux cas d'ISO détectés et validés permettront d'améliorer le gold-standard et la robustesse de l'algorithme.

Une interface de validation est en cours de développement pour permettre aux chirurgiens de visualiser les prédictions de l'algorithme afin de confirmer ou d'infirmer la présence d'une ISO chez un patient. Cette méthode semi-automatisée permettra de gagner du temps pour la déclaration des ISO, d'éviter un biais de mémorisation et de permettre une couverture temporelle plus large pour la surveillance.

Un travail similaire est en cours pour la détection des ISO en neurochirurgie et pour les poses de prothèses articulaires. La méthode de TAL est plus facilement généralisable à d'autres chirurgies et moins chronophage que la première approche de sélection manuelle des variables dans plusieurs sources de données.

Ce travail s'inscrit dans la démarche actuelle d'utilisation des outils informatiques et des algorithmes d'apprentissage automatique pour améliorer la prévention et la qualité des soins. Les algorithmes proposés pourraient entrer dans le cadre de la mission Spicmi qui a pour objectif de passer à un nouveau système de surveillance national reposant principalement sur l'utilisation des données des SIH. La possibilité d'utiliser le même algorithme pour l'ensemble des établissements reste à démontrer.

### Conclusion

Dans cet article, nous montrons la faisabilité de détecter automatiquement des infections du site opératoire à partir d'un entrepôt de données au CHU de Bordeaux. Un outil de validation semi-automatique pourra permettre un gain de temps aux équipes en charge de la surveillance des infections du site opératoire. Deux algorithmes ont été développés, le premier utilise l'ensemble des données disponibles tandis que le second utilise seulement les données en texte libre et est donc plus facilement transposable à un autre établissement. Les premiers résultats sont prometteurs avec de bonnes performances répondant aux exigences de sensibilité d'un outil de détection. Bien que les résultats soient meilleurs en utilisant l'ensemble des données, l'algorithme utilisant le texte libre seul fournit des résultats très proches. Ces résultats sont pour l'instant limités à la chirurgie du rachis et les algorithmes nécessitent d'être évalués et améliorés à partir de nouvelles données annotées.

### Références

**Adresse de correspondance**

marine.queroue@chu-bordeaux.fr